\documentclass[sigconf]{acmart}

\AtBeginDocument{%
  }

\setcopyright{acmcopyright}
\copyrightyear{2024}
\acmYear{2024}
\acmDOI{XXXXXXX.XXXXXXX}

\acmConference[ICBET 2024]{International Conference on Biomedical Engineering and Technology}{June 14--17,
  2024}{Seoul, South Korea}
\acmPrice{15.00}
\acmISBN{978-1-4503-XXXX-X/18/06}

\begin{document}

\title{Multi-Layer Dense Attention Decoder for Polyp Segmentation}

\author{Krushi Patel}
\affiliation{%
  \institution{The University of Kansas}
  \streetaddress{1450 Jayhawk Blvd}
  \city{Lawrence}
  \country{Kansas}}
\email{krushi92@ku.edu}

\author{Fengjun Li}
\affiliation{%
  \institution{The University of Kansas}
  \streetaddress{1450 Jayhawk Blvd}
  \city{Lawrence}
  \country{Kansas}}
\email{fli@ku.edu}

\author{Guanghui Wang}
\affiliation{%
  \institution{Toronto Metropolitan University}
  \city{Toronto}
  \state{ON}
  \country{Canada}}
  \email{wangcs@torontomu.ca}

\renewcommand{\shortauthors}{Patel et al.}

\begin{abstract}
Detecting and segmenting polyps is crucial for expediting the diagnosis of colon cancer. This is a challenging task due to the large variations of polyps in color, texture, and lighting conditions, along with subtle differences between the polyp and its surrounding area. Recently, vision Transformers have shown robust abilities in modeling global context for polyp segmentation. However, they face two major limitations: the inability to learn local relations among multi-level layers and inadequate feature aggregation in the decoder. To address these issues, we propose a novel decoder architecture aimed at hierarchically aggregating locally enhanced multi-level dense features. Specifically, we introduce a novel module named Dense Attention Gate (DAG), which adaptively fuses all previous layers' features to establish local feature relations among all layers. Furthermore, we propose a novel nested decoder architecture that hierarchically aggregates decoder features, thereby enhancing semantic features. We incorporate our novel dense decoder with the PVT backbone network and conduct evaluations on five polyp segmentation datasets: Kvasir, CVC-300, CVC-ColonDB, CVC-ClinicDB, and ETIS. Our experiments and comparisons with nine competing segmentation models demonstrate that the proposed architecture achieves state-of-the-art performance and outperforms the previous models on four datasets. The source code is available at: https://github.com/krushi1992/Dense-Decoder.
\end{abstract}




\maketitle

\section{Introduction}
Polyp segmentation plays a critical role in expediting the diagnosis of colorectal cancer, a disease recognized as one of the most prevalent cancers globally  \cite{mathur2020cancer}\cite{patel2021enhanced}\cite{silva2014toward}.  Detecting polyps at an early stage can significantly reduce the mortality rate. Colonoscopy is an effective technique for colorectal cancer (CRC) screening, identifying polyps that may lead to colon cancer. However, the challenge arises from the similarity in appearance between polyps and background pixels, making it difficult even for experienced clinicians to discern and potentially resulting in missed detections \cite{patel2020comparative}\cite{li2021colonoscopy}.  Moreover, polyps exhibit wide variations in size,  texture,  and color.  Therefore, an accurate and automated polyp segmentation method is imperative for the early detection of cancerous polyps, to decrease the mortality rate \cite{jia2019wireless}.

Convolutional neural networks (CNNs) have been extensively employed for polyp segmentation \cite{akbari2018polyp}\cite{brandao2017fully}\cite{fang2019selective}\cite{he2021sosd}\cite{jha2019resunet++}\cite{ronneberger2015u}\cite{zhou2018unet++}. 
 Specifically, various U-Net-shaped encoder-decoder-based segmentation networks \cite{jha2019resunet++} \cite{ronneberger2015u}\cite{zhou2018unet++} have demonstrated remarkable performance gains by generating high-resolution segmentation and aggregating multi-stage features through skip connections. However, they still struggle to establish long-range dependencies essential for accurate polyp segmentation. To address this limitation, some works incorporate attention-based modules into the segmentation architecture \cite{patel2021enhanced}\cite{patel2022fuzzynet}\cite{wei2021shallow}\cite{zhang2020adaptive}, leading to performance improvements. Nevertheless, these approaches still fall short of fully capturing long-range dependencies.

Recently, vision Transformer-based encoders \cite{dong2021polyp}\cite{zhang2021transfuse} have gained popularity and been successfully applied in polyp segmentation tasks due to their capability to capture long-range dependencies. Vision Transformers use an attention-based module to learn correlations among spatial patches, enabling them to capture the global context. However, vision Transformers are computationally expensive. To reduce the computational cost, hierarchical vision Transformers with modified attention modules have been proposed, such as PVT \cite{wang2021pyramid} and Swin Transformer \cite{liu2021swin}. These hierarchical vision Transformers have achieved significant performance gains in various computer vision-related tasks. However, we believe that Transformer-based segmentation networks fail to establish relationships among neighboring pixels.

Various segmentation networks such as polyp-PVT \cite{dong2021polyp} have attempted to address the aforementioned challenges by embedding convolution layers in the decoder network. However, these approaches primarily establish local relationships, leading to two main issues: (i) They do not fully exploit the multi-level features from previous layers during local feature modeling, which are crucial for identifying fine-grained clues. (ii) They lack hierarchical feature flows in the decoder to enhance the local feature relationships of global features progressively captured by the Transformer encoder. To resolve these issues, we propose a novel decoder network with dense connections and hierarchical feature flow. Specifically, we introduce a module called the dense attention gate, which considers all previous layers' decoder features via dense connections followed by an attention mechanism. Furthermore, we propose a novel multi-layer decoder to further refine the local features by hierarchically aggregating them, thereby improving feature flow.

The main contributions of this paper are summarized below:

\begin{enumerate}

\item \textbf{Dense Attention Gate:} 
We propose a novel module called the dense attention gate, which considers all the previous layers' encoding layer features to calculate spatial attention scores. These scores are then broadcastly multiplied by the current encoding layer features, rather than directly fusing the encoding features to the respective decoding layer.

\item \textbf{ Hierarchical multi layer decoder:} 
We introduce a hierarchical decoding layer to enhance the flow of features, refining local features through the horizontal extension of decoding layers. This involves utilizing the output of the previous decoding layers as inputs until a singular feature map is obtained.

\item \textbf{Improved performance:} Comprehensive experiments demonstrate that the integration of the dense attention gate module, along with the hierarchical multi-layer decoder, achieves superior performance across various polyp segmentation datasets and outperforms most other approaches. 


\end{enumerate}

\section{Related Work}
There have been various approaches proposed to segment the polyp from the colonoscopic images.  Based on the fundamental component used,  they mainly categorized into three categories: 

\textbf{Traditional Computer Vision Approach: } 
Before the emergence of neural networks, polyp segmentation predominantly relied on manually crafted features such as size, shape, texture, and color \cite{mamonov2014automated}. In the seminal study \cite{maghsoudi2017superpixel}, a simple linear iterative clustering superpixel method was proposed for polyp segmentation. However, these traditional approaches suffer from slow processing speeds and a high misdetection rate. This is mainly due to the limited representational capacity of hand-crafted features and the significant similarity between polyps and their surrounding areas.

\textbf{CNN-based Deep Learning Approach: }
Following the success of convolutional neural networks in various computer vision tasks, researchers have turned to CNN-based networks for polyp segmentation. This trend began with the work of \cite{akbari2018polyp}, where a modified version of Fully Convolutional Network (FCN) was employed to segment polyps, outperforming classical computer vision-based methods. Subsequently, U-Net \cite{ronneberger2015u} introduced a U-shaped encoder-decoder architecture, significantly improving performance. Building upon this success, various U-shaped models such as U-Net++ \cite{zhou2018unet++} and ResUNet++ \cite{jha2019resunet++} have been proposed, further enhancing polyp segmentation performance.

Although these networks achieved remarkable performance, they often struggle to differentiate between the polyp boundary and the surrounding area due to unfiltered encoded features. Furthermore, U-shaped encoder-decoder architectures typically employ a single-layer decoder, which we believe is insufficient for further improving decoded features. To address this limitation and enhance feature flow, we propose a horizontally extended multi-layer decoder to refine local features further.
To address the boundary issue, various attention-based models such as PraNet \cite{fan2020pranet}, ACSNet \cite{zhang2020adaptive}, Enhanced U-Net \cite{patel2021enhanced}, and SANet \cite{wei2021shallow} have been proposed. These models aim to suppress insignificant features and enhance important ones, thereby improving boundary prediction. However, none of them consider all the previous encoding layer features when calculating the attention score.

This paper proposes a novel module, named Dense Attention Gate, by taking into account all the previous layers' encoding features while calculating the attention score. This is essential for capturing multi-scale and fine-grained features, which can significantly improve boundary delineation.
\begin{figure*}[t]
   \includegraphics[width=1.0\linewidth]{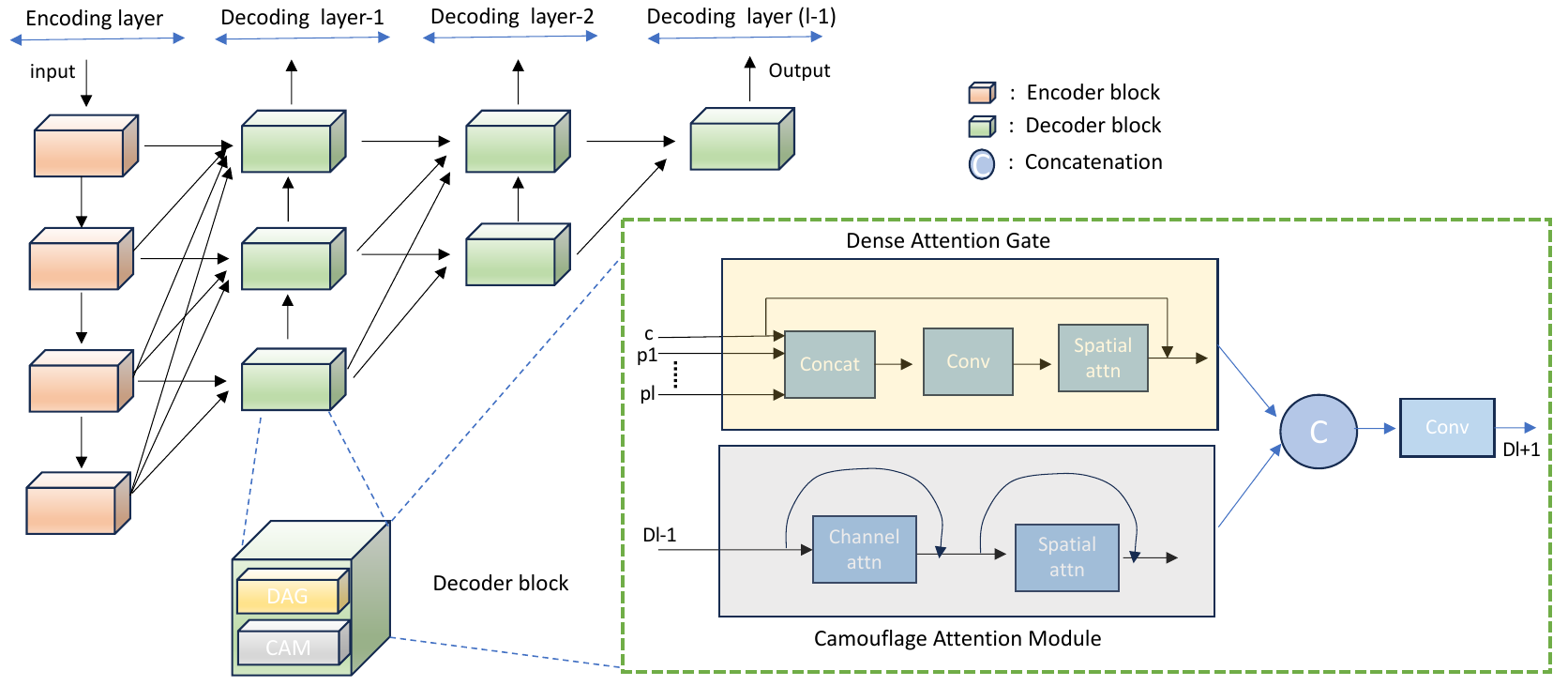} 
   \caption{The overall architecture of the multi-layer dense decoder-based polyp segmentation network. It comprises a PVT encoder represented by a collection of orange blocks. The dense decoder consists of multiple horizontally stacked decoding layers, each composed of a series of decoding blocks interconnected, denoted as green blocks. Each decoding block encompasses two modules: DAG (Dense Attention Gate) and CAM (Camouflage Attention Module). The block diagrams of both modules are depicted in the figure outlined within the green box.}
\label{fig:figure1}
\end{figure*} 

\textbf{Vision Transformer based Deep learning Approach: }
Vision Transformers have garnered significant attention and adoption in various computer vision tasks \cite{chen2023accumulated}\cite{wang2023pst}\cite{xiao2023edge}. Building upon their success, researchers have begun integrating them as backbone networks in polyp segmentation tasks, as demonstrated by models such as Polyp-PVT \cite{dong2021polyp}, TransFuse \cite{zhang2021transfuse}, and FuzzyNet \cite{patel2022fuzzynet}. This integration has led to notable performance improvements attributed to the establishment of long-range dependencies. However, despite these advancements, existing approaches still struggle to establish fully comprehensive neighborhood relationships between pixels.

In our work, we propose the integration of a dense attention gate mechanism to refine local features. Additionally, our design incorporates a hierarchical multi-layer decoder to further enhance local features within the global context generated by the Transformer-based encoder. This comprehensive approach aims to address the limitations of existing methods and further improve the accuracy of polyp segmentation.

\section{Method}
Our proposed network architecture consists of two main components as shown in Fig. \ref{fig:figure1}: the Transformer-based Encoder and the Dense Decoder. The model takes an RGB colonoscopic image as input and passes it through a series of attention layers within the PVT encoder. This process generates feature maps at each encoding layer's stage. Subsequently, these feature maps are fed into the corresponding stages of the decoding layers in our dense decoder. The dense decoder comprises a series of decoding layers stacked horizontally. Specifically, our model consists of three stacked decoding layers, each decoding layer consists of a series of decoding blocks composed of dense attention gates and camouflage attention modules arranged vertically. Below, we provide a detailed description of each component for clarity.

\subsection{Transformer -- Encoder }

In segmentation tasks, the encoder serves as the backbone of the network, generating fundamental multi-resolution features at each stage of refinement, denoted as ${f_{i}, i = 1,2,..4}$. It is considered the cornerstone of the model architecture, as the model heavily relies on the encoding features. While convolutional neural network-based encoding models are widely used for polyp segmentation tasks, they often struggle to generate global features essential for segmentation. Therefore, we employ a Transformer-based encoder, whose main component is the self-attention mechanism. This mechanism establishes long-range dependencies between all pixels and is capable of generating global features after each stage of processing.

\subsection{Multi-layer Dense Decoder}

To fully leverage the global features generated by the Transformer encoder, we introduce a novel decoder architecture called the Multi-layer Dense Decoder as shown in Fig. \ref{fig:figure1}. It consists of $l - 1$ decoding layers stacked together, with each decoding layer composed of a series of decoding blocks stacked vertically. Each decoding block comprises a dense attention gate followed by the camouflage attention module. Specifically, the main three components of the multi-layer dense decoder include Dense Attention Gate, Camouflage Attention Module,
and Multi-layer Decoder Design. The description of each component is as follows:
\begin{figure*}[tph]
   \includegraphics[width=1.0\linewidth]{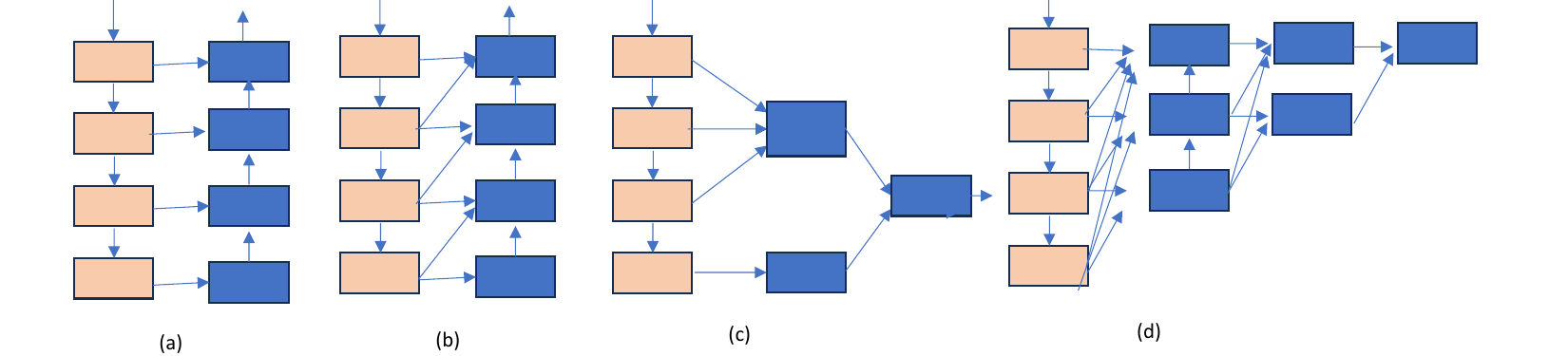} 
   \caption{Different types of decoding structures for the polyp segmentation. (a) U-shaped decoding structure \cite{ronneberger2015u}. (b) U-shaped decoding structure with previous layer integration strategy \cite{patel2021enhanced}. (c) Separate decoding for low-level and high-level features \cite{dong2021polyp}. (d) multi-layer dense decoder architecture proposed on our model.}
\label{fig:figure2}
\end{figure*} 

\textbf{Dense Attention Gate:}
In our approach, we replace the skip connection with the Dense Attention Gate, which considers all previous encoding block features when calculating the attention score. This differs from previously proposed attention modules, which typically only use current layer features. Specifically, we calculate the dense attention score by first concatenating the current encoding block features with all the previous encoding block features. Then, we apply a spatial attention block, which represents the significance of each pixel in the feature map. During the concatenation operation, we first upsample the previous layer features to match the size of the current layer feature map.
\begin{equation}
    C_i = Concat(E_{i},  ...  E_{l-i})
\end{equation}
\begin{equation}
    S_i = \sigma(Conv(C_{i}))
\end{equation}
\begin{equation}
    D_i =  E_{i} * S_{i}
\end{equation}
where $E_{i}$ is the current encoding block features and $l$ is the total number of blocks in the encoding layer.  Here we have $l=4$. $C_{i}$ is the concatenated features of encoding block $i$ and its previous layers.  $S_{i}$ is the spatial attention score calculated by applying a convolution operator followed by sigmoid activation.  $S_{i}$ is then multiplied by the current encoding block features which are then used as input to the Camouflage Attention Module.

\textbf{Camouflage Attention Module: }
To further enhance the distinction between polyp features and background elements, we incorporated the Camouflage Identification Module proposed in Polyp-PVT within each decoding block following the dense attention gate. In contrast to Polyp-PVT, where the CAM is applied only once on the low-level features, we integrate the CAM into each block of our decoder. 

The Camouflage Identification Module mainly consist of the channel attention mechanism $A_{c}(.)$ and spatial attention mechanism $A_{s}(.)$, which can be formulated as:
\begin{equation}
    D_{ci} = A_{s}(A_{c}(D_{i}))
\end{equation}
\begin{equation}
    A_{c}(D_{i}) = \sigma(\beta_{1}(P_{max}(D_{i})) +  \beta_{2}(P_{avg}(D_{i})))
\end{equation}
\begin{equation}
    A_{c}(D_{i}) = A_{c}(D_{i}) * D_{i}
\end{equation}
where $D_{i}$ is the output of the Dense attention gate.  $\sigma$ is the softmax activation.  $P_{max}$ and $P_{avg}$ denotes adaptive maximum pooling and adaptive average pooling respectively.  $\beta_{1}$ and $\beta_{2}$ are convolution operators of size $1 \times 1$ to reduce the channel dimension 16 times, followed by a ReLU layer and another $1 \times 1$ convolutional layer to recover the original channel dimension.  The spatial attention $A_{s}$ can be formulated as: 
\begin{equation}
    A_{s}(D_{i}) = \sigma(\alpha(Concat(R_{max}(D_{i}), R_{avg}(D_{i}))))
\end{equation}
\begin{equation}
    A_{s}(D_{i}) = A_{s}(D_{i}) * D_{i}
\end{equation}
where $R_{max}$ and $R_{avg}$ represent the maximum and average values obtained along the channel dimension, respectively.  $\alpha$ represent the convolution operation with $7 \times 7$ kernel and padding 3.

\textbf{Multi-layer Decoder Design: }
To enhance the differentiation between polyps and their background while refining local features, we expanded the decoding layers horizontally. A visual comparison between our decoder design and previous approaches is illustrated in Fig. \ref{fig:figure2}. Specifically, instead of a single decoding layer, we incorporated a total of \(l-1\) layers, where \(l\) represents the total number of encoding blocks. The output of one decoding layer serves as the input to the subsequent decoding layer within its respective decoding block. The design of decoding blocks in subsequent layers remains consistent, comprising dense attention gates and camouflage identification modules. This approach progressively enhances the local features within the global context generated by the Transformer encoder, facilitating the differentiation of polyps. At the conclusion of each decoding layer, we apply deep supervision loss. The final output is computed by aggregating the endpoint outputs from all the decoding layers.

\subsection{Loss Function}

We formulate the loss function as a combination of weighted IOU loss $L^{w}_{IOU}$ and weighted cross entropy loss $L^{w}_{BCE}$ \cite{qin2019basnet}\cite{wei2020f3net}.  We apply deep supervision at the end of each decoding layer. 
\begin{equation}
    L_{main} = L_{IOU} + L_{BCE}
\end{equation}
\begin{equation}
    L_{total} = \sum_{i=1}^{l-1} L_{main}(D_{i})
\end{equation}
where $l$ is the total number of encoding blocks. Our total number of decoding layers is dependent on the number of encoding blocks.  Specifically, in our case we have $l=4$, which lead to total number of decoding layers $l-1 = 3$.

\section{Experiments}
\subsection{Datasets and Models}

We conducted experiments on five publicly available polyp segmentation datasets:  Kvasir \cite{jha2020kvasir}, CVC-300 \cite{vazquez2017benchmark}, CVC-ColonDB \cite{tajbakhsh2015automated}, CVC-ClinicDB \cite{bernal2015wm}, and ETIS \cite{silva2014toward}. The ETIS dataset comprises 196 polyp images along with their corresponding ground truth masks. CVC-ClinicDB and CVC-300 consist of 612 and 300 images extracted from 29 and 13 colonoscopy video sequences, respectively. The CVC-ColonDB dataset is relatively small-scale, containing 380 images derived from 15 short colonoscopy sequences. The Kvasir dataset, on the other hand, is comparatively newer, consisting of 1000 polyp images. We compare our model with the following nine state-of-the-art models: PraNet \cite{fan2020pranet}, Enhanced U-Net \cite{patel2021enhanced}, ACSNet \cite{zhang2020adaptive}, MSEG \cite{huang2021hardnet}, SA- \cite{wei2021shallow}, TransFuse \cite{zhang2021transfuse}, Polyp-PVT \cite{dong2021polyp}, U-Net \cite{ronneberger2015u}, U-Net++ \cite{zhou2018unet++}, and ResU-Net++ \cite{jha2019resunet++}.

\begin{table}[t]
\caption{The results obtained on the CVC-ClinicDB and Kvasir datasets serve as indicators of the model's learning capability. The results reveal that our model surpasses other models by a substantial margin on the CVC-ClinicDB dataset, while achieving comparable performance on the Kvasir dataset. The results represent the average of three experiments. }
\begin{center}
\begin{tabular}{ l c c c c }
\toprule
Model & \multicolumn{2}{c}{CVC-ClinicDB}& \multicolumn{2}{c}{Kvasir}\\ 
\hline
 &mDice &mIoU&mDice & mIoU \\
\midrule
U-Net & 0.823 & 0.755& 0.818& 0.746\\
U-Net++ &0.794 &0.729&0.821 & 0.743\\
SFA & 0.700& 0.607& 0.723& 0.611\\
ACSNet &0.882 &0.826 &0.898& 0.838 \\
PraNet &0.899 &0.849& 0.898 &0.840 \\
EU-Net &0.902 &0.846 &0.908&0.854\\
SA-Net &0.916 &0.859 &0.904 & 0.847\\
TransFuse&0.918 &0.868 &0.918&0.868 \\
Polyp-PVT &0.937 &0.889&0.917&0.864  \\
Ours & \textbf{0.939}&\textbf{0.890}&\textbf{0.919}& \textbf{0.869} \\
\bottomrule

\end{tabular}
\label{tab:tab1}
\end{center}
\end{table}

\begin{table}[htb]
\caption{The results obtained on the CVC-ColonDB, ETIS, and CVC-300 datasets serve as indicators of the model's generalization capability. The results illustrate that our model outperforms other models by a substantial margin on the CVC-ColonDB and ETIS datasets, while achieving comparable performance on the CVC-300 dataset. The results represent the average of three experiments.}
\begin{center}
\begin{tabular}{ l c c c c c c}
\toprule
Model & \multicolumn{2}{c}{CVC-ColonDB}& \multicolumn{2}{c}{ETIS}& \multicolumn{2}{c}{CVC-300}\\ 
\hline
 &mDice &mIoU&mDice & mIoU&mDice&mIoU \\
\midrule
U-Net & 0.512 & 0.444& 0.398& 0.335& 0.710&0.627\\
U-Net++ &0.483 &0.410&0.401 & 0.344& 0.707&0.624\\
SFA & 0.469& 0.347& 0.297& 0.217&0.467&0.329 \\
ACSNet &0.716 &0.649 &0.578& 0.509&0.863&0.787 \\
PraNet &0.712 & 0.640& 0.628 &0.567&0.871&0.797 \\
EU-Net &0.756 &0.681 &0.687 &0.609& 0.837&0.765\\
SA-Net &0.753 &0.670 &0.750 & 0.654&0.888&0.815 \\
TransFuse&0.773 &0.696&0.733&0.659&\textbf{0.902}&\textbf{0.833} \\
Polyp-PVT &0.808 &0.727&0.787&0.706 &0.900&0.833 \\
Ours& \textbf{0.818}&\textbf{0.731} &\textbf{0.795}&\textbf{0.711} & 0.886 & 0.815 \\

\bottomrule

\end{tabular}
\label{tab:tab2}
\end{center}
\end{table}

\subsection{Evaluation Metric}
We employ the Dice similarity coefficient (DSC) and Intersection over Union (IOU) as our evaluation metrics, defined as follows:
\begin{equation}
    DSC(A,B) = \frac{2 \times (A \cap B)}{A \cup B}
\end{equation}
\begin{equation}
    IOU(A,B) = \frac{A \cap B}{A \cup B}
\end{equation}
where $A$ denotes the predicted set of pixels and $B$ is the ground truth of the image. $A$ denotes the predicted set of pixels and $B$ is the ground truth of the image.

\subsection{Implementation Details}
In our experimental setup, we adopt the training configurations consistent with those employed in Polyp-PVT. Our model is trained on one NVIDIA V100 GPU with a batch size of 16 using the Adam optimizer. We initialize the learning rate at 0.0001 and utilize multi-scale training, following the methodology outlined in PraNet instead of employing data augmentation techniques. The model initialization involves pre-trained weights trained on the ImageNet 1K dataset, with training extending across all layers. Network training spans 50 epochs for three iterations, and we report the average of the best results obtained.

\begin{table}[t]
\caption{ The results show the performance of the model with different numbers of decoding layers. }
\begin{center}
\begin{tabular}{ c c c c }
\toprule
Dataset & One Layer & Two Layer & Three Layer\\ 
\hline

Setting-1& 0.919 & 0.925 & 0.930\\
Setting-2 & 0.825&0.830 & 0.834\\

\bottomrule

\end{tabular}
\label{tab:tab3}
\end{center}
\end{table}

\subsection{Learning Ability}

\textbf{Setting:} 
We evaluate the learning capability of our model using the ClinicDB and Kvasir-Seg datasets. ClinicDB consists of 612 images extracted from 31 colonoscopy videos, while Kvasir-Seg contains a total of 1000 polyp images. Following the settings established by PraNet and Polyp-PVT, we partition 900 and 548 images from the ClinicDB and Kvasir-Seg datasets, respectively, for training purposes. The remaining 64 and 100 images are reserved for evaluation, serving as the test set.

\textbf{Results:}
Table \ref{tab:tab1} presents the performance metrics on the CVC-ClinicDB and Kvasir datasets, reflecting the learning efficacy of the model. In terms of the Dice score, our model demonstrates superiority over SFA, ACSNet, PraNet, EU-Net, SA-Net, TransFuse, and PolypPVT by 23.9\%, 5.7\%, 4\%, 3.7\%, 2.3\%, 2.1\%, and 0.2\%, respectively on the CVC-ClinicDB dataset.
Moreover, our model exhibits enhanced performance on the Kvasir dataset, surpassing SFA, ACSNet, PraNet, EU-Net, SA-Net, TransFuse, and Polyp-PVT by 19.6\%, 2.1\%, 2.1\%, 1.1\%, 1.5\%, 0.1\%, and 0.2\%, respectively, in terms of the Dice score. The notable achievements on both the CVC-ClinicDB and Kvasir datasets underscore the robust learning capabilities of our model.

\subsection{Generalization Ability}

\textbf{Setting:}
To assess the generalization capacity of the model, we utilize three previously unseen datasets: ETIS, ColonDB, and CVC-300. The ETIS dataset comprises a total of 190 images, while ColonDB consists of 380 images, and CVC-300 contains 60 images. These datasets encompass images sourced from various medical centers, implying that the training and testing sets are distinct, and the model has not encountered the test images during the training phase.

\textbf{Results:}
The performance evaluation of our model on the CVC-ColonDB, ETIS, and CVC-300 datasets is shown in Table \ref{tab:tab2}. The results indicate that our model achieves state-of-the-art performance on the CVC-ColonDB dataset, surpassing SFA, ACSNet, PraNet, EU-Net, SA-Net, TransFuse, and Polyp-PVT by 34.9\%, 10.2\%, 10.6\%, 6.2\%, 6.5\%, 4.5\%, and 1.0\%, respectively. Similarly, on the ETIS dataset, our model outperforms SFA, ACSNet, PraNet, EU-Net, SA-Net, TransFuse, and Polyp-PVT by 49.8\%, 21.7\%, 16.7\%, 10.8\%, 4.5\%, 6.2\%, and 0.8\%, respectively, in terms of the Dice score. On the CVC-300 dataset, our model surpasses SFA, ACSNet, PraNet, and EU-Net by 41.9\%, 2.3\%, 1.5\%, and 4.9\%, respectively, in terms of the Dice score, and achieves comparable results with Polyp-PVT. The consistent performance across all three datasets demonstrates the robust generalization ability of our model.

\subsection{Ablation Study}

\textbf{Effect of Multi-layer Decoder:}
We performed an ablation study to evaluate the impact of the number of decoding layers on the performance metrics of learning ability (setting-1) and generalization ability (setting-2). The results are shown in Table \ref{tab:tab3}. 
The results demonstrate that increasing the number of horizontal decoding layers leads to a corresponding improvement of 0.5\% and 1.06\% in accuracy on setting-1. This trend suggests that a higher number of decoders is positively correlated with increased accuracy. Furthermore, the performance on setting-2 shows a gradual enhancement as the number of decoding layers increases, with improvements of approximately 0.55\% and 1.0\% in terms of dice score. These observed enhancements in performance on both setting-1 and setting-2 affirm that augmenting the number of decoding layers effectively refines both local and global features.

\section{Conclusion}

This study has introduced a novel dense attention gate mechanism aimed at refining the local relationships across multi-level encoder features by incorporating all previous layer features during attention score computation. Additionally, a novel multi-layer decoder architecture has been developed to further augment semantic features. Integrating both of these new design modules into a PVT-based encoder, the proposed structure yields significant performance improvements in five public datasets for polyp segmentation. These results not only underscore the efficacy of our proposed methodology but also open new avenues for advancing the state-of-the-art in medical image analysis.

\section*{Acknowledgments}
The project is partly supported by the Natural Sciences and Engineering Research Council of Canada (NSERC).


\balance
\bibliographystyle{ACM-Reference-Format}
\bibliography{sample-base}

\end{document}